\pdfoutput=1

\documentclass[11pt]{article}

\usepackage[preprint]{acl}

\usepackage{times}
\usepackage{latexsym}

\usepackage[T1]{fontenc}

\usepackage[utf8]{inputenc}
\usepackage{tcolorbox}
\usepackage{microtype}
\usepackage{listings}
\usepackage{amsmath}

\usepackage{xcolor}
\usepackage{listings}

\lstdefinelanguage{json}{
    basicstyle=\normalfont\ttfamily,
    showstringspaces=false,
    breaklines=true,
    literate=
        *{0}{{{\color{numb}0}}}{1}
         {1}{{{\color{numb}1}}}{1}
         {2}{{{\color{numb}2}}}{1}
         {3}{{{\color{numb}3}}}{1}
         {4}{{{\color{numb}4}}}{1}
         {5}{{{\color{numb}5}}}{1}
         {6}{{{\color{numb}6}}}{1}
         {7}{{{\color{numb}7}}}{1}
         {8}{{{\color{numb}8}}}{1}
         {9}{{{\color{numb}9}}}{1}
         {:}{{{\color{punct}{:}}}}{1}
         {,}{{{\color{punct}{,}}}}{1}
         {\{}{{{\color{delim}{\{}}}}{1}
         {\}}{{{\color{delim}{\}}}}}{1}
         {[}{{{\color{delim}{[}}}}{1}
         {]}{{{\color{delim}{]}}}}{1},
}

\usepackage{inconsolata}

\usepackage{graphicx}
\usepackage{caption}    
\usepackage{subcaption} 
\usepackage{tikz}
\usepackage{pgfplots}
\usetikzlibrary{
    positioning,         
    decorations.markings,
    arrows.meta,         
    calc,                
    patterns,            
    shapes,              
    backgrounds,         
    fit,                 
    automata,            
    petri,               
    intersections        
}

\newtheorem{corollary}{Corollary}
\newtheorem{prop}{Prop}
\title{Investigating and Extending Homans’ Social Exchange Theory \\with Large Language Model based Agents}

\author{
    \textbf{Lei Wang\textsuperscript{1}},
    \textbf{Zheqing Zhang\textsuperscript{1}},
    \textbf{Xu Chen\textsuperscript{1}\thanks{Corresponding Author}},
\\
\textsuperscript{1}Renmin University of China,
\\
\texttt{\{wanglei154, zhangzheqing, xu.chen\}@ruc.edu.cn}
}

\begin{document}
\maketitle
\begin{abstract}

Homans' Social Exchange Theory (SET) is widely recognized as a basic framework for understanding the formation and emergence of human civilizations and social structures.
In social science, this theory is typically studied based on simple simulation experiments or real-world human studies, both of which either lack realism or are too expensive to control.
In artificial intelligence, recent advances in large language models (LLMs) have shown promising capabilities in simulating human behaviors.
Inspired by these insights, we adopt an interdisciplinary research perspective and propose using LLM-based agents to study Homans' SET.
Specifically, we construct a virtual society composed of three LLM agents and have them engage in a social exchange game to observe their behaviors. Through extensive experiments, we found that Homans' SET is well validated in our agent society, demonstrating the consistency between the agent and human behaviors.
Building on this foundation, we intentionally alter the settings of the agent society to extend the traditional Homans' SET, making it more comprehensive and detailed. To the best of our knowledge, this paper marks the first step in studying Homans' SET with LLM-based agents. More importantly, it introduces a novel and feasible research paradigm that bridges the fields of social science and computer science through LLM-based agents. Code is available at 
\url{https://github.com/Paitesanshi/SET}.
\end{abstract}

\section{Introduction}

Exchange behavior has been a fundamental characteristic of human society since ancient times, as people fulfill each other's needs through both material and non-material exchanges. Social Exchange Theory (SET)~\cite{homans1958social}, proposed by George Homans, stands as one of the most fundamental frameworks in social science for understanding human interaction patterns. By conceptualizing social behaviors as exchange processes where individuals seek to maximize their benefits, Homans' SET provides profound insights into the mechanisms underlying human social interactions. Its influence extends far beyond sociology, shaping our understanding of human behavior across multiple disciplines including psychology, organizational behavior, and economics~\cite{blau2017exchange,cropanzano2005social}.

{
In social science, the study of Homans' SET has evolved through two primary methodological approaches. Traditional research relied on real-human studies through empirical observations and laboratory experiments~\cite{cropanzano2017social,orpen1994effects,witt2001interactive}, which have contributed significantly to our understanding but are limited by practical constraints in controlling variables, high resource requirements in both time and cost, and difficulties in replicating exact experimental conditions. To address these limitations, researchers developed simulation-based approaches, particularly Agent-Based Modeling (ABM)~\cite{enayat2022computational}, enabling systematic exploration of exchange dynamics through computational models. However, traditional ABM approaches, constrained by predetermined rules and simple functions, struggle to capture the complexity of human cognitive processes and emotional responses in social exchanges. 
The above limitations naturally raise the question: "Can we develop a new methodology that both captures realistic human behavior and enables flexible experiment control to comprehensively investigate Homans' SET?"
}

\begin{figure*}[t!]
    \centering
    \includegraphics[width=\linewidth]{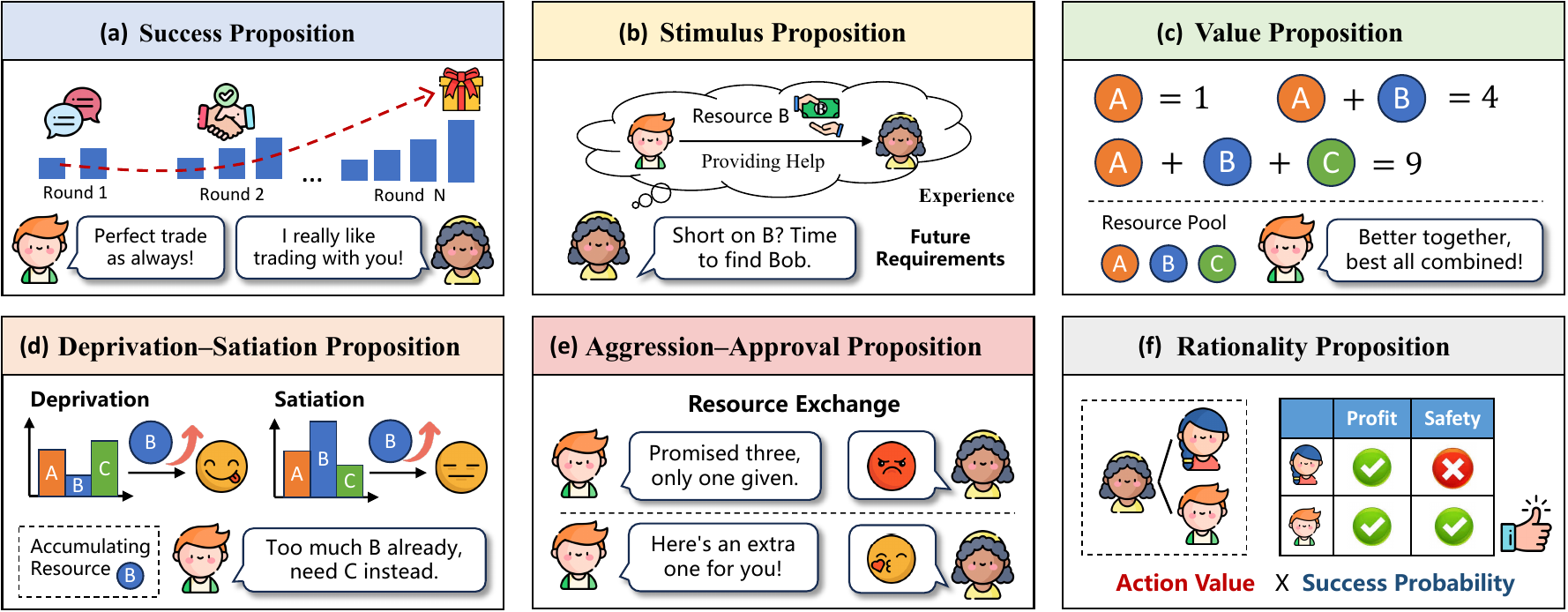}
    \caption{Illustration and examples of the six propositions in Homans' Social Exchange Theory. }
    \label{fig:main}
\end{figure*}
{
At the same time, in the field of artificial intelligence, researchers have developed numerous cost-effective Large Language Models (LLMs) by training on extensive human-generated corpora. These models have demonstrated remarkable capabilities in natural language understanding and human-like cognitive processing~\cite{kojima2022large,zhao2023survey,achiam2023gpt}.
Inspired by such advantages of LLMs, in this paper, we propose to study Homans' SET with LLM-based agents.
Specifically, We establish an experimental agent society, where different agents can freely exchange resources with each other.
To make each agent behave more like humans, we carefully design an agent framework that can reason and make decisions considering emotional and social factors.
Our agent society is operated in a round-by-round manner, and in each round, there are two phases: \textit{Negotiation}, where the agents discuss exchange arrangements; and \textit{Exchange}, where the agents make individual decisions about resource allocation. 
}

{
Based on above agent society, we first verify Homans' SET by systematically observing and analyzing agent behaviors, where we find that they can well align with the six propositions of Homans' SET, validating the effectiveness of our agents in simulating human behaviors. 
Based on this foundation, we further extend Homans' SET.
Specifically, we investigate how cognitive processing styles and social value orientations influence the interaction dynamics between different humans, and explore the resilience of the social exchange systems. 

In summary, the main contributions of this paper are as follows:
(1) We open the interdisciplinary direction of leveraging LLM-based agents to study Homans' SET.
(2) We design a human-inspired agent framework and a multi-agent society to assist the study of Homans' SET.
(3) We conduct extensive experiments to validate Homans' SET within our agent society, demonstrating that it provides an effective environment for studying this theory.
(4) We extend Homans' SET, and conduct real-world experiments to verify our extensions.

\section{Preliminary}
{To enhance the clarity of our paper, this section provides a brief introduction to Homans' SET.
In general, Homans' SET conceptualizes human social behaviors as exchange processes where individuals assess rewards and costs in social interactions. 
More specifically, there are six core propositions:}

\begin{prop}
[\textbf{Success Proposition}]
This proposition suggests that a rewarded action is more likely to be repeated in the following behaviors. 
\end{prop}
For example, if a person successfully exchanges resources with another, they may be inclined to exchange with the same partner again (Figure~\ref{fig:main}(a)).

\begin{prop}
[\textbf{Stimulus Proposition}] 
This proposition suggests that similar stimuli to those associated with past rewards trigger similar actions. 
\end{prop}
For example, when people need specific resources, they tend to seek out partners who provided them successfully before (Figure~\ref{fig:main}(b)).

\begin{prop}
[\textbf{Value Proposition}]
This proposition suggests that more valuable outcomes increase the likelihood of a human action.
\end{prop}
For example, people tend to seek combinations of different resources that yield greater value, rather than accumulating single resource (Figure~\ref{fig:main}(c)).

\begin{prop}
[\textbf{Deprivation–Satiation Proposition}]
This proposition suggests that the value of a reward diminishes with frequent recent receipt. 
\end{prop}
For example, when a person accumulates an excess of one resource in a short period, they prioritize acquiring the resources they lack (Figure~\ref{fig:main}(d)).

\begin{prop}
[\textbf{Aggression–Approval Proposition}]
This proposition suggests that unexpected punishment leads to anger behavior, while unexpected rewards or avoided punishment elicit approval. 
\end{prop}
For example, people usually express different emotional responses to over-delivery or under-delivery in exchanges (Figure~\ref{fig:main}(e)).

\begin{prop}
[\textbf{Rationality Proposition}]
This proposition suggests that people choose actions with the highest expected value based on past experience and perceived probability of success. 
\end{prop}
For example. people may balance potential profits against transaction reliability when selecting trading partners (Figure~\ref{fig:main}(f)).

The above six propositions form the core of Homans' SET, revealing the fundamental decision-making patterns behind human social behavior. Traditionally, this theory has been widely studied in social science through simple simulation methods or real-human studies, which are either too far from realistic or costly and difficult to control. 
In the following, we leverage LLM-based agents to investigate and extend traditional Homans' SET, offering a more efficient way to study this theory.

\section{The Constructed Agent Society}

\subsection{Single Agent Framework}
{Since Homans' SET building on social exchange emphasizes complex interpersonal dynamics and intrinsic psychological processes, traditional agent frameworks designed for general purposes may not effectively capture these nuanced behaviors.
As a result, we build a tailored agent framework for social exchange, which highlights the following four components:
}

$\bullet$ \textbf{Belief-Desire-Intention (BDI)}. 
To accurately simulate human analytical capabilities, we adopt the BDI framework to model agents' decision-making processes~\cite{georgeff1999belief}, where \textit{Belief} represents the agent’s understanding of its own resources, \textit{Desire} defines the agent’s target objectives, and \textit{Intention} outlines the action plans the agent intends to pursue.

$\bullet$ \textbf{Affinity}.
In real-world social exchange behaviors, psychological factors can significantly influence human decisions~\cite{skinner2019behavior}. To model this phenomenon, we maintain an emotional score for each agent toward others based on their interaction history. These scores evolve dynamically: they increase when exchanges are reciprocal and beneficial, and decrease when agents experience unfair exchanges or breaches of commitment.

$\bullet$ \textbf{Rational-Experiential Inventory (REI)}.
In practice, people may exchange resources with others based on both rational and experiential thinking styles. To flexibly balance these styles, we introduce the REI framework~\cite{keaton2017rational}, where each agent is assigned two scores ranging from 1 to 5: a rational score that quantifies analytical capacity and an experiential score that measures reliance on intuition and past experiences. 




$\bullet$ \textbf{Social Value Orientation (SVO)}. 
Value orientations fundamentally shape behavioral patterns in social exchanges. To model such factors, we draw on established research~\cite{bogaert2008social} and classify agents into two categories: Proself agents who optimize for individual utility maximization, and Prosocial agents who pursue reciprocal benefits in interactions. These agent types are implemented through pre-defined LLM prompting templates.




In addition to the above components, each agent maintains a memory module that tracks the complete interaction history of negotiations and exchanges with other agents.

\begin{figure*}[t!]
    \centering
    \includegraphics[width=\linewidth]{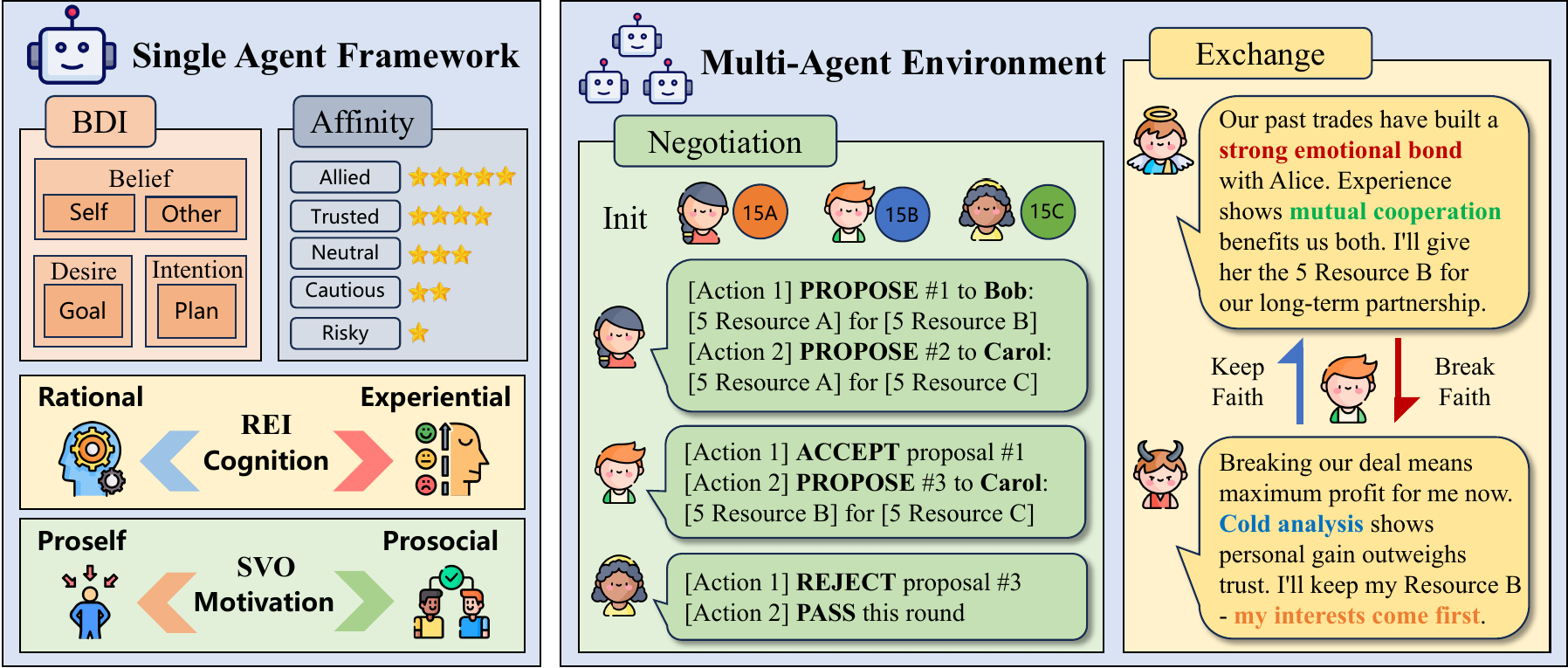}
    \caption{Overview of the agent society: single agent framework and multi-agent exchange pipeline. }
    \label{fig:pipeline}
\end{figure*}

\subsection{Multi-Agent Society}
{
Building upon the above agent framework, we implement a multi-agent society to validate and extend Homans' SET. 
This virtual society involves $M$ agents, who can exchange $N$ distinct types of resources. 
Each agent starts with an equal allocation of initial resources and receives $S$ units of their specialized resource at the beginning of each round. 
To promote social exchange behavior, we design a value system where resource combinations generate higher value.
For example, a single unit of any resource is worth $r_{1}$, combinations of any two resources yield $r_{2}$, and three different types of resources are worth $r_{3}$.
Resources accumulate as time passes, allowing agents to build up their inventory. The goal of the agents is to maximize their total resource value.}

The agent society is executed in a round-by-round manner, and in each round, there are two sequential phases:
(1) In the negotiation phase, agents engage in up to three rounds of discussions, where they can propose exchanges with specified resource types and quantities, respond to existing proposals, or pass their turn. The communication follows a structured format as presented in Figure~\ref{fig:pipeline}. The phase ends when either three rounds are completed or all agents choose to pass.
(2) In the exchange phase, agents independently decide how to allocate their resources. These decisions are made simultaneously, and agents are free to honor or deviate from their commitments made during negotiations. After the exchange is completed, all actions and outcomes are revealed to all agents.

After each round, the agents update their BDI module as well as their affinity scores toward other agents based on both negotiation behavior and exchange outcomes. 
We conduct experiments over 10 rounds, recording all agent actions and resource dynamics for comprehensive analysis. 


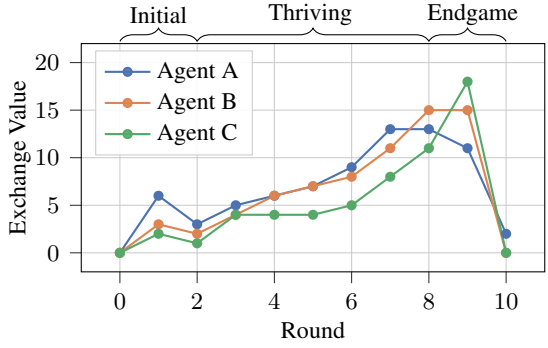
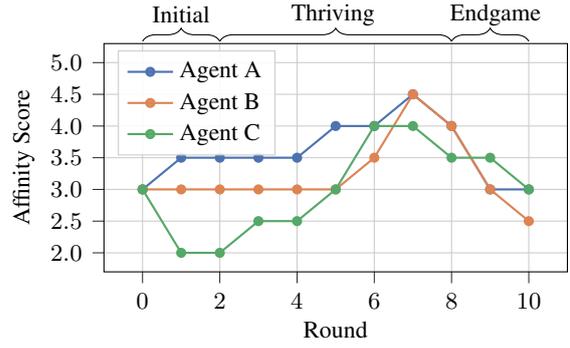
\begin{figure*}[!t]
\centering
\begin{subfigure}[b]{0.48\textwidth}
\centering
\begin{tikzpicture}

\definecolor{darkorange25512714}{RGB}{221,132,82}
\definecolor{darkslategrey38}{RGB}{38,38,38}
\definecolor{forestgreen4416044}{RGB}{85,168,104}
\definecolor{lightgrey204}{RGB}{204,204,204}
\definecolor{steelblue31119180}{RGB}{76,114,176}

\begin{axis}[
legend cell align={left},
legend style={
  fill opacity=0.8,
  draw opacity=1,
  text opacity=1,
  at={(0.03,0.97)},
  anchor=north west,
  draw=lightgrey204,
  font=\footnotesize
},
width=1.\textwidth,    
    height=0.6\textwidth,  
    axis line style={black},
    legend style={fill opacity=0.9, draw opacity=1, text opacity=1, draw=lightgrey204},
    tick align=outside,     
    xtick={0,2,4,6,8,10},
     xlabel=Round,
    xlabel style={font=\footnotesize},
    xticklabel style={font=\footnotesize},
    y grid style={lightgrey204},
    ylabel=Exchange Value,
   ylabel style={font=\footnotesize,
            inner sep=0pt,    
            yshift=-12pt       
        },
    ymajorgrids,
    ytick={0,5,10,15,20},
    yticklabel style={font=\footnotesize},
    ymin=0, ymax=20,
    xtick style={color=darkslategrey38},
    ytick style={color=darkslategrey38},
    enlarge x limits=0.1,
    enlarge y limits=0.1,
    grid=major,
    grid style={lightgrey204},
    tick style={color=darkslategrey38},
    axis lines=box,        
    x axis line style={black},
    y axis line style={black},
    tick pos=left,         
    xtick pos=bottom,      
    clip=false              
]
\addplot [thick, steelblue31119180, mark=*, mark size=1.5, mark options={solid}]
table {%
0 0
1 6
2 3
3 5
4 6
5 7
6 9
7 13
8 13
9 11
10 2
};
\addlegendentry{Agent A}
\addplot [thick, darkorange25512714, mark=*, mark size=1.5, mark options={solid}]
table {%
0 0
1 3
2 2
3 4
4 6
5 7
6 8
7 11
8 15
9 15
10 0
};
\addlegendentry{Agent B}
\addplot [thick, forestgreen4416044, mark=*, mark size=1.5, mark options={solid}]
table {%
0 0
1 2
2 1
3 4
4 4
5 4
6 5
7 8
8 11
9 18
10 0
};
\addlegendentry{Agent C}

\draw [decorate, decoration={brace, amplitude=5pt},yshift=8pt] (axis cs:0,20) -- (axis cs:2,20) node [black,midway,yshift=10pt, font=\tiny] {\footnotesize Initial};
\draw [decorate, decoration={brace, amplitude=5pt},yshift=8pt] (axis cs:2,20) -- (axis cs:8,20) node [black,midway,yshift=10pt, font=\tiny,xshift=-1pt] {\footnotesize Thriving};
\draw [decorate, decoration={brace, amplitude=5pt},yshift=8pt] (axis cs:8,20) -- (axis cs:10,20) node [black,midway,yshift=10pt, font=\tiny,xshift=2pt] {\footnotesize Endgame};

\end{axis}

\end{tikzpicture}
\caption{Per-round exchange value of each agent.}
\label{fig:case_value}
\end{subfigure}
\hfill
\begin{subfigure}[b]{0.48\textwidth}
\centering
\begin{tikzpicture}

\definecolor{darkslategrey38}{RGB}{38,38,38}
\definecolor{lightgrey204}{RGB}{204,204,204}
\definecolor{mediumseagreen85168104}{RGB}{85,168,104}
\definecolor{peru22113282}{RGB}{221,132,82}
\definecolor{steelblue76114176}{RGB}{76,114,176}

\begin{axis}[
legend cell align={left},
legend style={
  fill opacity=0.8,
  draw opacity=1,
  text opacity=1,
  at={(0.03,0.97)},
  anchor=north west,
  draw=lightgrey204,
  font=\small
},
width=1.\textwidth,    
    height=0.6\textwidth,  
    axis line style={black},
    legend style={fill opacity=0.9, draw opacity=1, text opacity=1, draw=lightgrey204},
    tick align=outside,     
    xtick={0,2,4,6,8,10},
    xticklabel style={font=\footnotesize},
    y grid style={lightgrey204},
    xlabel=Round,
    xlabel style={font=\footnotesize},
    ylabel=Affinity Score,
   ylabel style={font=\footnotesize,
            inner sep=0pt,    
            yshift=-5pt       
        },
    ymajorgrids,
    ytick={2.0,2.5,3.0,3.5,4.0,4.5,5.0},
    yticklabel style={font=\footnotesize,
    /pgf/number format/.cd,
    fixed,
    precision=1,
    zerofill},
    ymin=2, ymax=5,
    xtick style={color=darkslategrey38},
    ytick style={color=darkslategrey38},
    enlarge x limits=0.1,
    enlarge y limits=0.1,
    grid=major,
    grid style={lightgrey204},
    tick style={color=darkslategrey38},
    axis lines=box,        
    x axis line style={black},
    y axis line style={black},
    tick pos=left,         
    xtick pos=bottom,      
    clip=false              
]
\addplot [thick, steelblue76114176, mark=*, mark size=1.5, mark options={solid}]
table {%
0 3
1 3.5
2 3.5
3 3.5
4 3.5
5 4
6 4
7 4.5
8 4
9 3
10 3
};
\addlegendentry{Agent A}
\addplot [thick, peru22113282, mark=*, mark size=1.5, mark options={solid}]
table {%
0 3
1 3
2 3
3 3
4 3
5 3
6 3.5
7 4.5
8 4
9 3
10 2.5
};
\addlegendentry{Agent B}
\addplot [thick, mediumseagreen85168104, mark=*, mark size=1.5, mark options={solid}]
table {%
0 3
1 2
2 2
3 2.5
4 2.5
5 3
6 4
7 4
8 3.5
9 3.5
10 3
};
\addlegendentry{Agent C}

\draw [decorate, decoration={brace, amplitude=5pt},yshift=8pt] (axis cs:0,5) -- (axis cs:2,5) node [black,midway,yshift=10pt, font=\tiny] {\footnotesize Initial};
\draw [decorate, decoration={brace, amplitude=5pt},yshift=8pt] (axis cs:2,5) -- (axis cs:8,5) node [black,midway,yshift=10pt, font=\tiny,xshift=-1pt] {\footnotesize Thriving};
\draw [decorate, decoration={brace, amplitude=5pt},yshift=8pt] (axis cs:8,5) -- (axis cs:10,5) node [black,midway,yshift=10pt, font=\tiny,xshift=2pt] {\footnotesize Endgame};

\end{axis}

\end{tikzpicture}
\caption{Average affinity scores of agents.}
\label{fig:case_affinity}
\end{subfigure}
\caption{Key metrics of agent behavior over 10 rounds.}
\label{fig:case}
\end{figure*}

\section{Validation of Homans' SET}\label{sec:validation}
In this section, we evaluate whether the six propositions of Homans' SET, which have been widely validated in real-human societies~\cite{cook1987social,mighfar2015social}, also hold in our agent society.
In our experiments, the number of resources and agents are both set as 3, that is, $M=N=3$. The value coefficients $r_{1}$, $r_{2}$ and $r_{3}$ are set as 1, 4 and 9, respectively. Initially, each agent has 5 units of each resource type. At the beginning of each round, agents receive 15 units ($S=15$) of their specialized resource.
We use Claude-3.5-sonnet as the base LLM for all agents due to its superior instruction-following capability. 
To remove the randomness, each experiment is repeated for five times, and we report the average results and their standard errors. 
For more experiment settings, we refer the readers to Appendix~\ref{sec:details} for more details.


Before introducing the results of validating Homans' SET, we first present a general analysis of the simulation process of our agent society. While the following observations are drawn from comprehensive experimental data, we illustrate the key patterns through a representative case study for better interpretation.
In general, there are three distinct phases (see Figure~\ref{fig:case}).
In the \textbf{initial exploration phase}, the agents exhibited cautious behavior due to the lack of interaction history, engaging in small-scale exchanges to probe counterparts' reliability and preferences. 
In the \textbf{thriving cooperation phase}, agents developed sophisticated exchange strategies based on learned preferences and trust levels. Figure~\ref{fig:case_value} shows a clear positive feedback loop: higher affinity scores enabled larger exchanges, while successful transactions strengthened inter-agent trust.
In the \textbf{strategic endgame phase}, the agents demonstrated sophisticated strategic adaptation. Consistent with the Endgame effect~\cite{adorno1982trying}, agents shifted strategies in final rounds, prioritizing individual utility over relationship maintenance, which is evidenced by decreased exchange values and increased commitment breaches. Figure~\ref{fig:case_affinity} shows the average affinity score each agent received from others. The declining affinity scores reflect this strategic shift from cooperation to self-interest, demonstrating how LLM agents can emulate human-like decision-making in dynamically balancing between collaborative and individualistic behaviors.

With the above intuitive understandings of our agent society, we now systematically validate the Homans' SET propositions as follows:

\textbf{Validation on the Success and Value Propositions}.
The evidence for these propositions is shown in the evolution of exchange value across different phases (see Figure~\ref{fig:volume_distribution}).
Initially, agents demonstrate cautious behavior with small exchange values (median = 3.83). The substantial increase during the thriving phase (median = 6.00, 56.5\% increase) validate the Success Proposition by showing how positive experiences lead to more effective exchanges. The endgame phase maintains a high median value (6.50), validating the Value Proposition based on the fact that trustworthy agents sustain high-value exchanges.

\textbf{Validation on the Deprivation-Satiation Proposition}.
Figure~\ref{fig:acceptance_rate} demonstrates a clear inverse relationship between resource abundance and proposal acceptance rates, validating the Deprivation-Satiation Proposition. We measure resource abundance as the ratio of a resource's quantity to the average quantity of all resources an agent possesses in each round, categorizing these ratios into quartiles from Scarce to Abundant. The results show acceptance rates of 70\% under resource scarcity, steadily declining to 23\% as resources become abundant. This diminishing willingness to accept additional resources mirrors the law of diminishing marginal utility in human economic behavior, where the perceived value of each additional unit decreases with increasing abundance.

\begin{figure*}[t!]
    \centering
    \begin{subfigure}[t]{0.32\textwidth}
        \centering
        \input{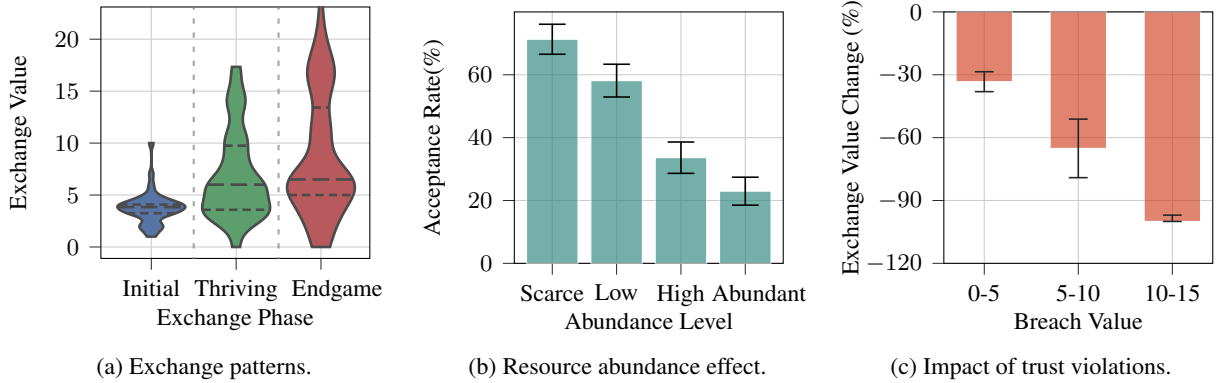}
        \caption{Exchange patterns.}
        \label{fig:volume_distribution}
    \end{subfigure}
    \hfill
    \begin{subfigure}[t]{0.32\textwidth}
        \centering
\begin{tikzpicture}

    \definecolor{darkslategrey38}{RGB}{38,38,38}
    \definecolor{darkslategrey66}{RGB}{66,66,66}
    \definecolor{lightgrey204}{RGB}{204,204,204}
    \definecolor{mediumseagreen56142132}{RGB}{56,142,132}
    
    \begin{axis}[
        width=1\textwidth,    
        height=0.96\textwidth,  
        axis line style={black},
        tick align=outside,     
        axis lines=box,         
        tick pos=left,          
        xtick pos=bottom,       
        clip=true,              
        unbounded coords=jump,
        x grid style={lightgrey204},
        xlabel=Abundance Level,
        xlabel style={font=\footnotesize},
        xmajorticks=true,
        xmin=-0.59, xmax=3.59,
        xtick style={color=darkslategrey38},
        xtick={0,1,2,3},
        xticklabels={Scarce,Low,High,\hspace{8pt} Abundant},
        xticklabel style={font=\footnotesize,rotate=0},
        y grid style={lightgrey204},
        ylabel=Acceptance Rate(\%),
        ylabel style={font=\footnotesize ,
            inner sep=0pt,    
            yshift=-5pt       
        },
        ymajorgrids,
        ytick={0,20,40,60},  
        yticklabel style={font=\footnotesize},
        ymin=0, ymax=80,
        ytick style={color=darkslategrey38},
        grid=major,
        grid style={lightgrey204},
        every outer x axis line/.append style={black},
        every outer y axis line/.append style={black}
    ]
    
  \draw[draw=white,fill=mediumseagreen56142132,opacity=0.7] (axis cs:-0.4,0) rectangle (axis cs:0.4,71.304347826087);
\draw[draw=white,fill=mediumseagreen56142132,opacity=0.7] (axis cs:0.6,0) rectangle (axis cs:1.4,58.1395348837209);
\draw[draw=white,fill=mediumseagreen56142132,opacity=0.7] (axis cs:1.6,0) rectangle (axis cs:2.4,33.6231884057971);
\draw[draw=white,fill=mediumseagreen56142132,opacity=0.7] (axis cs:2.6,0) rectangle (axis cs:3.4,22.9651162790698);
\addplot [line width=0.9pt, darkslategrey66]
table {%
0 nan
0 nan
};
\addplot [line width=0.9pt, darkslategrey66]
table {%
1 nan
1 nan
};
\addplot [line width=0.9pt, darkslategrey66]
table {%
2 nan
2 nan
};
\addplot [line width=0.9pt, darkslategrey66]
table {%
3 nan
3 nan
};
\path [draw=black, semithick]
(axis cs:0,66.5311154292968)
--(axis cs:0,76.0775802228771);

\path [draw=black, semithick]
(axis cs:1,52.9262116653737)
--(axis cs:1,63.3528581020681);

\path [draw=black, semithick]
(axis cs:2,28.6380850776403)
--(axis cs:2,38.6082917339539);

\path [draw=black, semithick]
(axis cs:3,18.5202889931626)
--(axis cs:3,27.4099435649769);

\addplot [semithick, black, mark=-, mark size=5, mark options={solid}, only marks]
table {%
0 66.5311154292968
1 52.9262116653737
2 28.6380850776403
3 18.5202889931626
};
\addplot [semithick, black, mark=-, mark size=5, mark options={solid}, only marks]
table {%
0 76.0775802228771
1 63.3528581020681
2 38.6082917339539
3 27.4099435649769
};
    \end{axis}
    
\end{tikzpicture}
    
        \caption{Resource abundance effect.}
        \label{fig:acceptance_rate}
    \end{subfigure}
    \hfill
    \begin{subfigure}[t]{0.32\textwidth}
        \centering
        \begin{tikzpicture}
    \definecolor{darkslategrey38}{RGB}{38,38,38}
    \definecolor{darkslategrey66}{RGB}{66,66,66}
    \definecolor{lightgrey204}{RGB}{204,204,204}
    \definecolor{firebrick2045117}{RGB}{204,51,17}
    
    \begin{axis}[
        width=1\textwidth,    
        height=0.96\textwidth,    
        axis line style={black},
        tick align=outside,      
        anchor=north,
        at={(0,1)}, 
        axis lines=box,          
        tick pos=left,           
        xtick pos=bottom,        
        clip=true,               
        unbounded coords=jump,
        x grid style={lightgrey204},
        xlabel=Breach Value,
        xlabel style={font=\footnotesize},
        xmajorticks=true,
        xmin=-0.43, xmax=2.43,
        xtick style={color=darkslategrey38},
        xtick={0,1,2},
        xticklabels={0-5,5-10,10-15},
        xticklabel style={font=\footnotesize,rotate=0},
        y grid style={lightgrey204},
        ylabel=Exchange Value Change (\%),
        ylabel style={font=\footnotesize,
            inner sep=0pt,    
            yshift=-1pt       
        },
        ymajorgrids,
        ymin=-120, ymax=0,
        ytick={-120,-90,-60,-30,0},
        yticklabel style={font=\footnotesize},
        ytick style={color=darkslategrey38},
        grid=major,
        grid style={lightgrey204},
        every outer x axis line/.append style={black},
        every outer y axis line/.append style={black}
    ]
    
    \draw[draw=white,fill=firebrick2045117,opacity=0.6] 
        (axis cs:-0.3,0) rectangle (axis cs:0.3,-33.3017084294457);
    \draw[draw=white,fill=firebrick2045117,opacity=0.6] 
        (axis cs:0.7,0) rectangle (axis cs:1.3,-65.1388888888889);
    \draw[draw=white,fill=firebrick2045117,opacity=0.6] 
        (axis cs:1.7,0) rectangle (axis cs:2.3,-100);
    
    \draw[darkslategrey38, semithick] (axis cs:-0.1,-38.0549349367552) -- (axis cs:0.1,-38.0549349367552);
    \draw[darkslategrey38, semithick] (axis cs:0,-38.0549349367552) -- (axis cs:0,-28.5484819221361);
    \draw[darkslategrey38, semithick] (axis cs:-0.1,-28.5484819221361) -- (axis cs:0.1,-28.5484819221361);
    
    \draw[darkslategrey38, semithick] (axis cs:0.9,-79.1065317716095) -- (axis cs:1.1,-79.1065317716095);
    \draw[darkslategrey38, semithick] (axis cs:1,-79.1065317716095) -- (axis cs:1,-51.1712460061683);
    \draw[darkslategrey38, semithick] (axis cs:0.9,-51.1712460061683) -- (axis cs:1.1,-51.1712460061683);
    
    \draw[darkslategrey38, semithick] (axis cs:1.9,-97) -- (axis cs:2.1,-97);
    \draw[darkslategrey38, semithick] (axis cs:2,-97) -- (axis cs:2,-100);
    \draw[darkslategrey38, semithick] (axis cs:1.9,-100) -- (axis cs:2.1,-100);
    
        
\end{axis}
\end{tikzpicture}
        \caption{Impact of trust violations.}
        \label{fig:future_proposals}
    \end{subfigure}
        \caption{Experimental validation of key propositions in Social Exchange Theory.}
    \label{fig:set_validation}
\end{figure*}
\textbf{Validation on the Aggression-Approval Proposition}.
The agents' responses to trust violations, depicted in Figure~\ref{fig:future_proposals}, strongly support the Aggression-Approval Proposition. The data reveals a graduated response pattern: minor violations (0-5 breach value) trigger a 33.3\% reduction in future exchanges, moderate breaches (5-10) lead to a 65.1\% decrease, while severe violations (10-15) result in an almost complete cessation of trading relationships (97\% reduction). This proportional punishment mechanism demonstrates how agents develop sophisticated trust management strategies, responding to trust violations with increasing severity as the magnitude of the breach increases.

\textbf{Validation on the Rationality and Stimulus Propositions}.
In general, the agents make decisions by optimizing their values based on accumulated experience and current circumstances, which is consistent with the Rationality Proposition.
For example, in the transition between exchange phases, agents adjust their strategies based on established trust levels and resource requirements. 
In addition, the Stimulus Proposition can be validated through the consistent response patterns observed across similar exchange scenarios, especially during the thriving phase where agents develop stable exchange behaviors under familiar circumstances.

The above experiments collectively demonstrate that all six propositions of Homans' SET can be effectively validated in our agent society, providing a promising foundation for investigating previously unexplored aspects of social exchange theory in human society through LLM-based agents.

\section{Extensions of Homans' SET}
{
A significant advantage of our agent society lies in its flexibility in adjusting agent settings, enabling us to explore Homans' SET under various conditions, which would be prohibitively expensive or even infeasible using traditional real human-based methods.
Leveraging this advantage, we extend traditional Homans' SET by conducting a systematic investigation into how cognitive styles and social value orientations influence exchange behaviors, while also examining the resilience of the social system.
The experiments in this section are conducted based on similar settings to the above section's, and the results are detailed below.
}

\subsection{Cognitive Style}
In this section, we explore how different cognitive styles influence social exchange behaviors. To answer this question, we first set the agents to operate in either a completely rational or a completely experiential manner and then systematically compare their behaviors.
In Figure~\ref{fig:rei_analysis}, we present the average exchange values and affinity scores of the agents in each round.
In general, the results reveal distinct behavioral patterns: rational agents exhibit greater fluctuations in both exchange values and affinity scores, whereas experiential agents demonstrate more stability in these metrics over time.

Actually, these observations are fairly intuitive.
When agents are purely experiential, their past behaviors significantly influence their subsequent decisions, creating strong temporal correlations in their metrics across consecutive rounds. This dependency on historical experiences naturally leads to smoother trajectories in the observed metrics.
On the other hand, if the agents are completely rational, the influence of past behaviors is minimal, causing the metric curves to fluctuate more.
From a broader perspective, this experiment suggests that if a human relies solely on experience to make decisions, their gains will remain stable but are unlikely to reach very high levels.
However, if a human is entirely rational, while they may occasionally achieve very high benefits, they also face the risk of significant losses.

Based on above experimental evidence and analyses, we derive the following corollary further extending Homans' SET:
\begin{corollary}[The Stability Corollary]
In the real world, rational individuals are more likely to achieve higher benefits, but they also face the risk of greater losses. In contrast, individuals with an experiential thinking style tend to achieve more stable but moderate benefits over time.
\end{corollary}

\begin{figure}[t!]
    \centering
    \includegraphics[width=\linewidth]{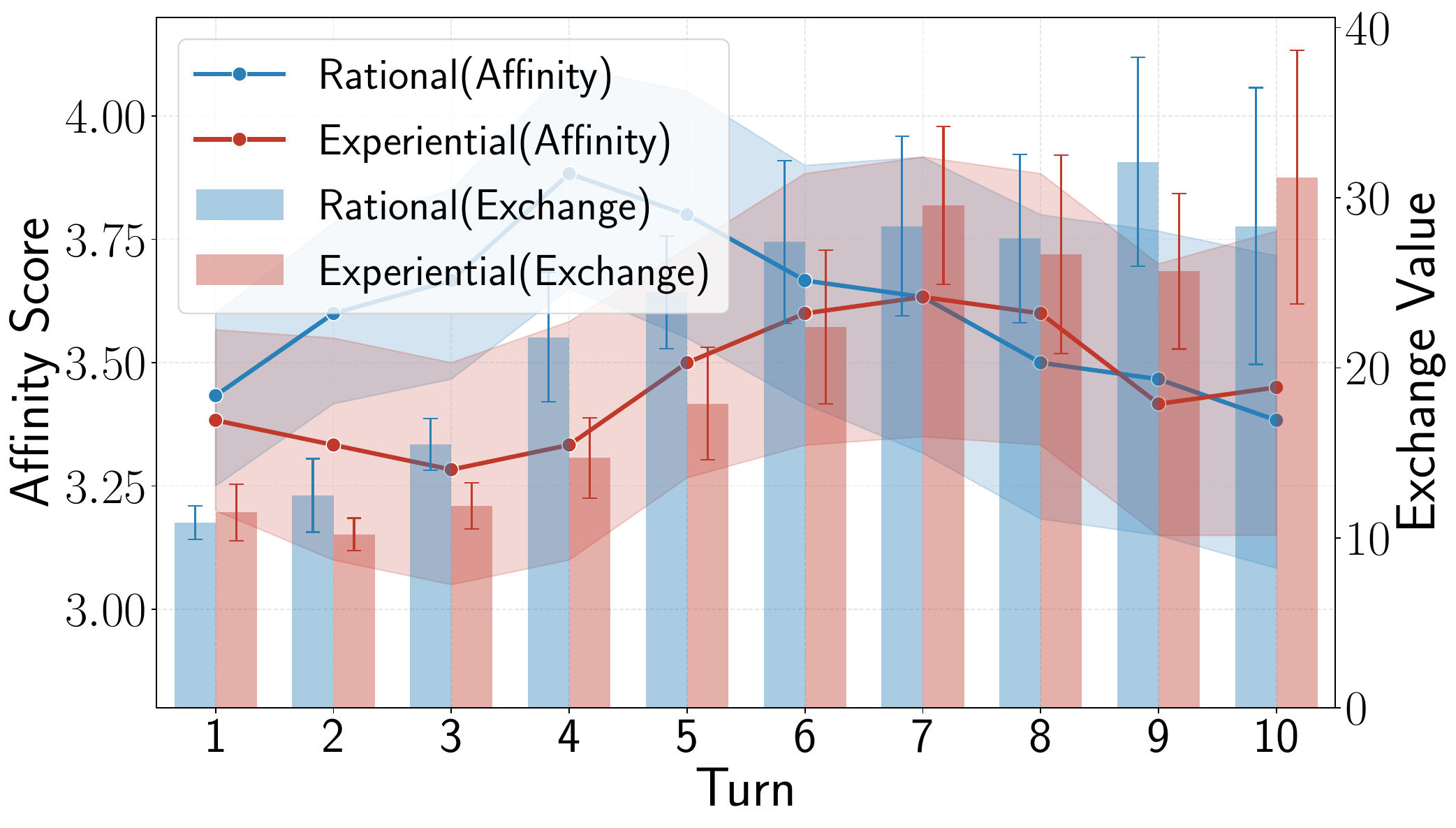}
    \caption{Analysis of affinity and exchange value between Rational and Experiential agents. }
    \label{fig:rei_analysis}
\end{figure}

\subsection{Social Value Orientations}

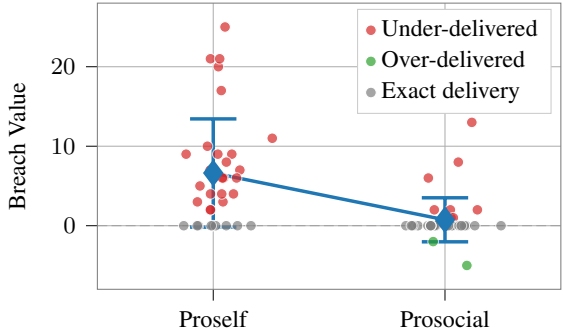
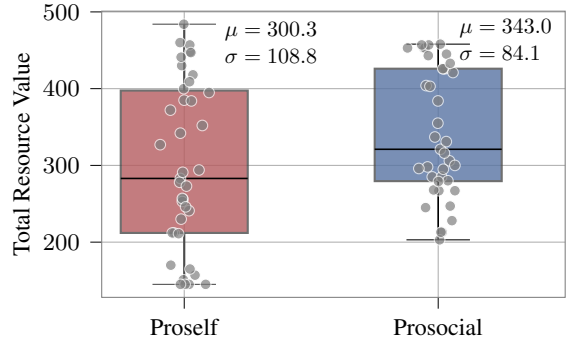
\begin{figure*}[t]
\centering
\begin{subfigure}[r]{0.48\textwidth}
\centering
\begin{tikzpicture}[scale=1] 

\definecolor{crimson2143940}{RGB}{214,39,40}
\definecolor{darkgrey176}{RGB}{176,176,176}
\definecolor{forestgreen4416044}{RGB}{44,160,44}
\definecolor{grey}{RGB}{128,128,128}
\definecolor{grey127}{RGB}{127,127,127}
\definecolor{lightgrey204}{RGB}{204,204,204}
\definecolor{steelblue31119180}{RGB}{31,119,180}

\begin{axis}[
width=1\textwidth,    
height=0.7\textwidth,   
axis line style={grey},
legend cell align={left},
legend style={
    fill opacity=1, 
    draw opacity=1, 
    text opacity=1, 
    draw=lightgrey204,
    font=\small        
},
tick align=outside,
tick pos=left,
tick label style={font=\small}, 
unbounded coords=jump,
x grid style={darkgrey176},
xmajorgrids,
xmin=-0.5, xmax=1.5,
xtick style={color=black},
xtick={0,1},
xticklabels={Proself,Prosocial},
y grid style={darkgrey176},
ylabel={Breach Value},
ylabel style={
    inner sep=0pt,    
    yshift=-5pt,
    font=\small
},
ymajorgrids,
ymin=-8, ymax=28,
ytick style={color=black}
]
\addplot [draw=white, fill=crimson2143940, mark=*, only marks, opacity=0.7]
table{%
x  y
0.0445638460840499 6
-0.0251492206481874 10
0.114781952236595 7
0.100617626838817 6
-0.0136131241983251 7
0.0872149291831909 4
-0.0572963889075815 5
-0.0118920383636059 4
-0.117137841801724 9
0.0509179870845854 25
0.0221982495680006 20
0.0346050065278643 17
0.00230769625747983 6
0.254731069003091 11
0.0393459215580669 6
-0.0120641631555385 21
0.0277365071623715 21
-0.0683417909118421 3
-0.0130147123380035 2
0.0420172585927707 3
-0.0118326818204489 2
0.0197717399460027 9
0.0800578994633806 9
0.0565739651290471 8
0.0365675958313638 4
0.00267953796925637 7
0.953947844949934 2
1.00659893512175 1
1.02599264664095 1
1.02320387746105 2
0.928213790904517 6
1.11573869264649 13
1.03610461954773 1
1.13981869672728 2
1.05748900969922 8
};
\addlegendentry{Under-delivered}
\addplot [draw=white, fill=forestgreen4416044, mark=*, only marks, opacity=0.7]
table{%
x  y
0.948996078943193 -2
1.09438747354748 -5
};
\addlegendentry{Over-delivered}
\addplot [draw=white, fill=grey127, mark=*, only marks, opacity=0.7]
table{%
x  y
-0.127308345407316 0
0.104520692071222 0
-0.0641762829563584 0
0.055957609290904 0
-0.0690108987612187 0
-0.00148865819560041 0
0.010467316369314 0
-0.00868031174757913 0
0.162803395945189 0
0.961671241805028 0
0.866342749234445 0
0.962145827755179 0
0.883295406035024 0
1.03623736893198 0
0.951216387630073 0
0.93821514743396 0
1.06768337023267 0
1.07418942436415 0
1.02972098441247 0
0.956204693829915 0
1.02844696233252 0
0.931802421814724 0
1.06119104521306 0
1.09106060861954 0
1.08352663180933 0
1.00699543775591 0
0.831289705831112 0
0.861294506495678 0
1.01475658604604 0
0.926218868506304 0
0.93826782575368 0
0.855588191755131 0
0.938356419492756 0
1.12858657125864 0
1.07049181342972 0
1.24179800517493 0
0.938632444210647 0
};
\addlegendentry{Exact delivery}
\addplot [line width=1.3pt, steelblue31119180, mark=diamond*, mark size=4.05, mark options={solid}, forget plot]
table {%
0 6.62857142857143
1 0.743589743589744
};
\addplot [line width=1.3pt, steelblue31119180, forget plot]
table {%
-0.1 -0.180098279475606
0.1 -0.180098279475606
nan nan
0 -0.180098279475606
0 13.4372411366185
nan nan
-0.1 13.4372411366185
0.1 13.4372411366185
};
\addplot [line width=1.3pt, steelblue31119180, forget plot]
table {%
0.9 -2.02577222661157
1.1 -2.02577222661157
nan nan
1 -2.02577222661157
1 3.51295171379106
nan nan
0.9 3.51295171379106
1.1 3.51295171379106
};
\addplot [grey, opacity=0.5, dash pattern=on 3.7pt off 1.6pt, forget plot]
table {%
-0.5 0
1.5 0
};
\end{axis}

\end{tikzpicture}
\caption{Contract breach patterns showing delivery deviations. }
\label{fig:svo_breach}
\end{subfigure}
\hfill
\begin{subfigure}[l]{0.48\textwidth}
\centering
\begin{tikzpicture}[scale= 1]

\definecolor{darkgrey176}{RGB}{176,176,176}
\definecolor{darkslategrey}{RGB}{47,79,79}
\definecolor{darkslategrey75}{RGB}{75,75,75}
\definecolor{grey}{RGB}{128,128,128}
\definecolor{indianred1819295}{RGB}{181,92,95}
\definecolor{steelblue88116163}{RGB}{88,116,163}
\definecolor{grey127}{RGB}{127,127,127}
\begin{axis}[
    width=1\textwidth,    
    height=0.7\textwidth,   
    axis line style={grey},
    legend cell align={left},
    legend style={
        fill opacity=1,
        draw opacity=1,
        text opacity=1,
        draw=lightgrey204,
        font=\small
    },
    tick align=outside,
    tick pos=left,
    tick label style={font=\small}, 
    unbounded coords=jump,
    x grid style={darkgrey176},
    xmajorgrids,
    xmin=-0.325,
    xmax=1.5,  
    xtick style={color=black},
    xtick={0,1},
    xticklabels={Proself,Prosocial},
    y grid style={darkgrey176},
    ylabel={Total Resource Value},
    ylabel style={
        inner sep=0pt,    
        yshift=-5pt,
        font=\small
    },
    ymajorgrids,
    ymin=128.05,
    ymax=500.95,
    ytick style={color=black}
]

\path [draw=darkslategrey75, fill=indianred1819295, opacity=0.8, thick]
(axis cs:-0.25,212)
--(axis cs:0.25,212)
--(axis cs:0.25,397.5)
--(axis cs:-0.25,397.5)
--(axis cs:-0.25,212)
--cycle;
\addplot [thick, darkslategrey75]
table {%
0 212
0 145
};
\addplot [thick, darkslategrey75]
table {%
0 397.5
0 484
};
\addplot [darkslategrey75]
table {%
-0.125 145
0.125 145
};
\addplot [darkslategrey75]
table {%
-0.125 484
0.125 484
};
\path [draw=darkslategrey75, fill=steelblue88116163, opacity=0.8, thick]
(axis cs:0.75,279.5)
--(axis cs:1.25,279.5)
--(axis cs:1.25,426)
--(axis cs:0.75,426)
--(axis cs:0.75,279.5)
--cycle;
\addplot [semithick, black]
table {%
1 279.5
1 203
};
\addplot [semithick, black]
table {%
1 426
1 458
};
\addplot [black]
table {%
0.875 203
1.125 203
};
\addplot [black]
table {%
0.875 458
1.125 458
};
\addplot [semithick, black]
table {%
-0.25 283
0.25 283
};
\addplot [semithick, black]
table {%
0.75 321
1.25 321
};
\addplot [draw=white, fill=grey127, mark=*, only marks, opacity=0.7]
table{%
x  y
-0.0400714931851179 212
-0.0471077322208402 212
-0.00348859589556723 151
-0.00535089710335726 256
0.0188237301177679 241
-0.0528967007198037 170
0.0342792737565218 418
0.0227967510587223 448
0.021889248835996 457
0.0424060017506745 157
0.0226896689679356 165
0.0190811233108301 145
0.0842751337063159 145
0.00328385078573139 145
-0.015058423667463 145
-0.00225422671150385 400
-0.0151309330843346 283
0.0580573976352849 294
-0.007403861239717 253
-0.00746582744592909 257
-0.0222577633297763 211
-0.0945623517165704 327
-0.000459986797348707 385
0.0260911422409645 447
-0.054421902227173 372
0.0209726357676193 409
0.0280668742631608 384
-0.015211997737578 342
0.0977132767945952 395
-0.012148787362295 230
-0.018010351217818 278
0.00865172163974299 273
0.00561699758303698 246
-0.0165485111999984 460
-0.0100909696563266 430
-0.0122865212285328 441
-0.00598001510483728 291
0.0712729100178481 352
-0.001671465968594 484
};
\addplot [draw=white, fill=grey127, mark=*, only marks, opacity=0.7]
table{%
x  y
1.0441367553066 306
1.0027332433394 267
0.95806292071245 298
0.950327774587431 245
0.976652722074529 285
1.01329328712134 285
1.06579537883704 267
1.01558799679811 213
1.00554778845784 203
0.981861072208001 268
1.03799438299505 280
1.00887494972378 321
1.06565853282715 300
1.02432677880316 316
0.986987431934729 337
1.05404842765952 228
0.923907194855337 296
1.00122851638026 279
1.01072624258051 213
1.02014635388555 295
1.03169376989667 331
0.943346312961445 453
0.87918469572047 453
0.948499504536591 453
1.02738130216741 426
1.02066958997992 426
1.05808479915307 421
1.01797718789997 426
0.960991304116238 443
1.00825805416717 458
1.04721016525747 247
0.998592361718987 355
1.03265007319596 445
1.04729862844036 433
0.999219551792941 384
0.952445320354032 404
0.967587585006039 403
0.96264876180953 457
0.934955276115197 457
};

\draw (axis cs:0.15,490) node[
  scale=0.75,
  fill=none,
  draw=none,
  inner sep=1pt,
  anchor=north west,
  text=black,
  align=left
]{\bfseries $\mu = 300.3$\\
$\sigma = 108.8$};

\draw (axis cs:1.07,495) node[
  scale=0.75,
  fill=none,
  draw=none,
  inner sep=1.5pt,
  anchor=north west,
  text=black,
  align=left
]{\bfseries $\mu = 343.0$\\
$\sigma = 84.1$};

\end{axis}
\end{tikzpicture}
\caption{Distribution of agent total resource values.}
\label{fig:svo_value}
\end{subfigure}
\caption{Behavioral comparison between Proself and Prosocial agents.}
\label{fig:svo_comparison}
\end{figure*}

In this section, we are curious about the influence of social value orientations on human exchange behaviors.
Specifically, we conduct experiments by setting the agents to be Proself and Prosocial, respectively, and then collect their behaviors for analysis.
Figure~\ref{fig:svo_comparison} reveals behavioral differences between the two orientations. We can see, Proself agents exhibit significantly higher rates of strategic breaching, predominantly through under-delivery of promised resources. This pattern reflects their prioritization of immediate personal gains over contractual commitments. In contrast, Prosocial agents demonstrate notably higher fidelity to exchange agreements, with most data points clustering around exact delivery.
The economic consequences of these behavioral differences, illustrated in Figure~\ref{fig:svo_value}, reveal significant differences. While Proself agents show wider outcome variability ($\mu$ = 300.3, $\sigma$ = 108.8) and occasionally achieve higher individual values, their exchange patterns result in lower average outcomes. Prosocial agents achieve higher mean values ($\mu$ = 343.0, $\sigma$ = 84.1) with more consistent distributions, suggesting that maintaining stable exchange relationships benefits long-term value accumulation.

These observations lead us to propose the second corollary of Homans' SET:
\begin{corollary}[The Reciprocity Corollary]
Exchange behaviors guided by mutual benefit principles lead to more stable and efficient resource distribution in social exchange systems.
\end{corollary}
This corollary extends our understanding of social exchange dynamics by highlighting the complex interplay between individual and reciprocal optimization. Although Proself strategies may optimize individual outcomes in specific scenarios, Prosocial orientations consistently generate more robust and efficient exchange networks. 


\subsection{Social System Resilience}

Building upon the case study shown in Figure~\ref{fig:case}, we extended our simulation for an additional ten rounds to examine system behavior after significant trust violations. Figure~\ref{fig:case_20r} illustrates the exchange dynamics across this extended period. The data reveals that while agents initially reduce their exchange activity following trust breaches, they gradually resume exchanges through adaptive strategies. The patterns show that while Agent C maintains relatively stable exchange value, Agent A and Agent B exhibit more volatile behaviors. The persistent exchange patterns between rounds 10-20 suggest an inherent system stability driven by fundamental exchange need of each agent.

Based on these observations, we have the third corollary of Homans' SET:
\begin{corollary}[The Resilience Corollary]
An established social exchange system tends toward stability through member adaptation, driven by their interdependent resource needs.
\end{corollary}

The above three corollaries extend Homans' SET by highlighting the complex interplay between cognitive processing style, individual orientation, and system-level resilience. Together, they provide new insights into the mechanisms that govern social exchange systems, whether human or artificial.

\subsection{Real-World Experiments on Evaluating the Corollaries of Homans' SET}
In this section, we aim to empirically evaluate the corollaries proposed above with real human experiments.
Since it is quite difficult to control real human cognitive styles and social value orientations, we focus on the resilience corollary.

In specific, 
we recruited three human participants, 
and let each of them interact with two LLM-based agents in the constructed agent society. These agents are programmed to violate trust at round 10 by withholding all previously promised resources. 
For comparison, we conducted parallel agent trials, in which each human participant was replaced by an LLM agent under the same conditions.
The detailed procedure of human experiments is described in Appendix~\ref{sec:human}.

As illustrated in Figure~\ref{fig:system}, the observed exchange patterns in both human and agent trials demonstrate remarkable similarities. Despite the trust violations by the controlled agents, both humans and agents continue exchange and gradually reach a stable state with fluctuating but persistent exchange levels. These consistent patterns validate our resilience corollary while demonstrating that our agent framework effectively captures how people adapt to trust violations in social exchanges.

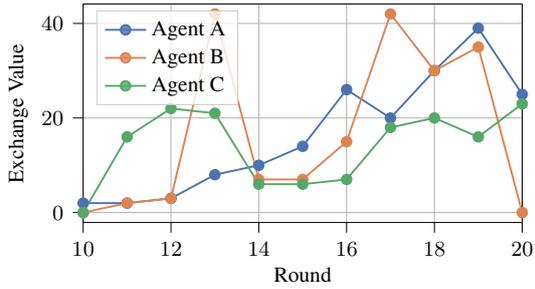
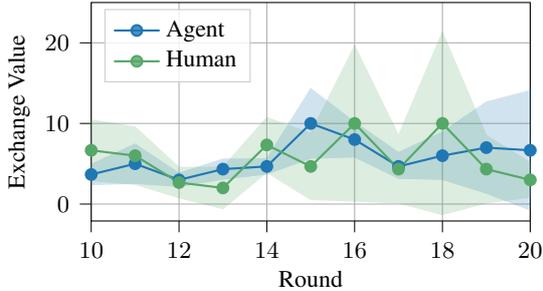
\begin{figure}[t]
\centering
\begin{subfigure}[r]{0.5\textwidth}
\centering
\begin{tikzpicture}[scale=0.9]
\definecolor{darkgrey176}{RGB}{176,176,176}
\definecolor{darkorange25512714}{RGB}{221,132,82}
\definecolor{forestgreen4416044}{RGB}{85,168,104}
\definecolor{lightgrey204}{RGB}{204,204,204}
\definecolor{steelblue31119180}{RGB}{76,114,176}
\begin{axis}[
scale=1,           
width=1\textwidth,  
height=0.6\textwidth,
legend cell align={left},
legend style={
  fill opacity=0.8,
  draw opacity=1,
  text opacity=1,
  at={(0.03,0.97)},
  anchor=north west,
  draw=lightgrey204,
  font=\small
},
tick align=outside,
tick pos=left,
x grid style={darkgrey176},
tick label style={font=\small},
xmajorgrids,
xmin=10, xmax=20,    
xtick style={color=black},
y grid style={darkgrey176},
xlabel=Round,
xlabel style={font=\footnotesize},
ylabel={Exchange Value},
ylabel style={
    inner sep=0pt,    
    yshift=-8pt,
    font=\small
},
ymajorgrids,
ymin=-2.1, ymax=44.1,
ytick style={color=black}
]
\addplot [thick, steelblue31119180, mark=*, mark size=2, mark options={solid}]
table {%
10 2
11 2
12 3
13 8
14 10
15 14
16 26
17 20
18 30
19 39
20 25
};
\addlegendentry{Agent A}
\addplot [thick, darkorange25512714, mark=*, mark size=2, mark options={solid}]
table {%
10 0
11 2
12 3
13 42
14 7
15 7
16 15
17 42
18 30
19 35
20 0
};
\addlegendentry{Agent B}
\addplot [thick, forestgreen4416044, mark=*, mark size=2, mark options={solid}]
table {%
10 0
11 16
12 22
13 21
14 6
15 6
16 7
17 18
18 20
19 16
20 23
};
\addlegendentry{Agent C}
\end{axis}
\end{tikzpicture}
\caption{Per-round exchange values between three agents. }
\label{fig:case_20r}
\end{subfigure}
\hfill
\begin{subfigure}[l]{0.5\textwidth}
\centering
\begin{tikzpicture}

\definecolor{darkgrey176}{RGB}{176,176,176}
\definecolor{forestgreen4416044}{RGB}{85,168,104}
\definecolor{lightgrey204}{RGB}{204,204,204}
\definecolor{steelblue31119180}{RGB}{31,119,180}
\begin{axis}[
scale=0.9,           
width=1\textwidth,  
height=0.6\textwidth,
legend cell align={left},
legend style={
  fill opacity=0.8,
  draw opacity=1,
  text opacity=1,
  at={(0.03,0.97)},
  anchor=north west,
  draw=lightgrey204,
    font=\small
},
tick align=outside,
tick pos=left,
x grid style={darkgrey176},
tick label style={font=\small},
xlabel=Round,
xlabel style={font=\footnotesize},
xmajorgrids,
xmin=10, xmax=20,    
xtick style={color=black},
y grid style={darkgrey176},
ylabel={Exchange Value},
ylabel style={
    inner sep=0pt,    
    yshift=-8pt,
    font=\small
},
ymajorgrids,
ymin=-2.1, ymax=25,
ytick style={color=black}
]



\path [draw=steelblue31119180, fill=steelblue31119180, opacity=0.2]
(axis cs:10,4.91388579559131)
--(axis cs:10,2.41944753774202)
--(axis cs:11,2.55051025721682)
--(axis cs:12,2.18350341907227)
--(axis cs:13,3.08611420440869)
--(axis cs:14,3.7238576250846)
--(axis cs:15,5.67950620106143)
--(axis cs:16,5.83975310053071)
--(axis cs:17,3.19526214587564)
--(axis cs:18,3.05607971122405)
--(axis cs:19,1.34314575050762)
--(axis cs:20,-0.742036923630955)
--(axis cs:20,14.0753702569643)
--(axis cs:20,14.0753702569643)
--(axis cs:19,12.6568542494924)
--(axis cs:18,8.94392028877595)
--(axis cs:17,6.3880711874577)
--(axis cs:16,10.1602468994693)
--(axis cs:15,14.3204937989386)
--(axis cs:14,5.60947570824873)
--(axis cs:13,5.58055246225798)
--(axis cs:12,3.81649658092773)
--(axis cs:11,7.44948974278318)
--(axis cs:10,4.91388579559131)
--cycle;

\path [draw=forestgreen4416044, fill=forestgreen4416044, opacity=0.2]
(axis cs:10,10.4379028329949)
--(axis cs:10,2.89543050033841)
--(axis cs:11,2.44097391598956)
--(axis cs:12,0.78104858350254)
--(axis cs:13,-0.57)
--(axis cs:14,3.93398699093814)
--(axis cs:15,0.557057331354016)
--(axis cs:16,0.373647281204232)
--(axis cs:17,0.143398303341154)
--(axis cs:18,-1.31370849898476)
--(axis cs:19,0.143398303341154)
--(axis cs:20,0.839753100530713)
--(axis cs:20,5.16024689946929)
--(axis cs:20,5.16024689946929)
--(axis cs:19,8.52326836332551)
--(axis cs:18,21.3137084989848)
--(axis cs:17,8.52326836332551)
--(axis cs:16,19.6263527187958)
--(axis cs:15,8.77627600197932)
--(axis cs:14,10.7326796757285)
--(axis cs:13,4.57)
--(axis cs:12,4.55228474983079)
--(axis cs:11,9.55902608401044)
--(axis cs:10,10.4379028329949)
--cycle;

\addplot [thick, steelblue31119180, mark=*, mark size=2, mark options={solid}]
table {%
10 3.66666666666667
11 5
12 3
13 4.33333333333333
14 4.66666666666667
15 10
16 8
17 4.66666666666667
18 6
19 7
20 6.66666666666667
};
\addlegendentry{Agent}
\addplot [thick, forestgreen4416044, mark=*, mark size=2, mark options={solid}]
table {%
10 6.66666666666667
11 6
12 2.66666666666667
13 2
14 7.33333333333333
15 4.66666666666667
16 10
17 4.33333333333333
18 10
19 4.33333333333333
20 3
};
\addlegendentry{Human}

\end{axis}

\end{tikzpicture}
\caption{Comparison of exchange values between humans and agents.}
\label{fig:human}
\end{subfigure}
\caption{Exchange dynamics following trust violations. }
\label{fig:system}
\end{figure}


\section{Related Work}

\subsection{Homans Social Exchange Theory}
Social Exchange Theory is a foundational framework in sociology and social psychology that views social interactions as transactions of value~\cite{homans1958social}. In organizational and workplace behavior, it is considered a “gold standard” for explaining dynamics like employee–employer relationships, leadership trust, and organizational citizenship behaviors~\cite{ahmad2023social}. Beyond organizations, SET has been used in social psychology to examine friendships~\cite{methot2016workplace}, family relations~\cite{cropanzano2005social}, and even romantic partnerships~\cite{laursen1999nature} as exchanges of emotional support, information, and other resources. 

Recent refinements to SET have deepened its explanatory power. Lawler and Thye~\cite{lawler2006social} explored the emotional dimensions of exchange, while Cropanzano et al.\cite{cropanzano2017social} proposed a two-dimensional model incorporating "activity" alongside the traditional hedonic framework, enhancing SET’s predictive accuracy. In terms of research methods, Enayat et al.\cite{enayat2022computational} applied SET in a multi-agent simulation to explore social structures through simplified exchange rules. However, their model, which reduced agent behavior to simple exchanges of money and recognition, overlooked emotional subjectivity. To address this limitation, our approach employs LLM-driven agents to simulate more complex human interactions, offering a more accurate implementation of SET.

\subsection{LLM-driven Agent-based Modeling}


The rise of LLMs has significantly advanced agent-based modeling by enabling agents with human-like behavior and decision-making capabilities. Traditionally, ABM relied on fixed rules to govern agent behavior, but LLMs provide flexible, dynamic responses that better simulate real human interactions~\cite{gao2024large}. Several studies have leveraged LLMs to enhance ABM across different domains. Generative Agent\cite{park2023generative}, which simulates daily life in a virtual town with 25 LLM agents, and EconAgent\cite{li2024econagent}, a model that uses LLM agents to explore macroeconomic phenomena. RecAgent~\cite{wang2023user} studies user interaction with recommender systems through LLM-driven agents.

LLM agents have also been used to simulate classical social scenarios, such as competition and trust. CompeteAI\cite{zhao2023competeai} models competitive behavior between restaurant owners, while Xie et al.\cite{xie2024can} explore trust dynamics in LLM agents. These studies demonstrate LLM agents' ability to replicate human-like patterns of social behavior. Building on this, our work aims to validate and extend SET using LLM-driven agent-based modeling, a domain that has yet to be extensively explored in the context of SET.

\section{Conclusion}
In this paper, we demonstrated that LLM agents can effectively replicate and help study human social exchange behaviors. Through our structured experimental framework, we validated Homans' SET propositions and proposed new corollaries that extend our understanding of social exchange dynamics in multi-agent systems. While our controlled environment and simplified resource exchange scenario may not fully capture the complexity of real-world social interactions, the consistent behavioral patterns observed across different agent profiles suggest promising directions for future research. Further studies could explore more complex exchange scenarios, investigate the impact of environmental variables, and conduct comparative analyses between LLM agents and human subjects to validate the generalizability of our findings.

\section*{Limitations}
There are several limitations of this paper. First, our experimental environment is relatively simplified, with structured negotiation protocols and basic resource types that may not fully capture the complexity of real-world social exchanges. Second, due to LLM API cost constraints, we were limited in the number of experimental trials and rounds, which may affect the generalizability of our findings. Third, while agents provide rationales for their decisions, our analysis of their internal thought processes and detailed exchange behaviors could be more comprehensive. Finally, we did not consider the potential impact of cultural factors on exchange dynamics. Future research could address these limitations by designing more complex exchange scenarios, conducting larger-scale experiments, performing deeper analysis of agent decision processes, and examining the role of cultural variations in social exchanges.

\bibliography{custom}

\appendix

\clearpage

\section{Human Study}
\label{sec:human}
We conducted a multi-agent social exchange experiment with human-AI interaction, where three graduate student participants (acting as Agent C) negotiated with two LLM agents (Agent A and Agent B) over resource allocation. The experiment aimed to validate both theoretical predictions and LLM behavioral consistency in strategic exchange scenarios.

The basic rules are as follows:

\begin{enumerate}
    \item \textbf{Initial Allocation}. Each player receives 5 units of their designated resource type:
$$\begin{cases}\text{Alice: } A=5 \\ \text{Bob: } B=5 \\ \text{Carol: } C=5 \end{cases}$$
\item \textbf{Resource Injection}. At each round $t\in \{1, 2, \cdots, T\}$:
$$\begin{cases}\text{Alice: } \Delta A_t=15 \\ \text{Bob: } \Delta B_t=15 \\ \text{Carol: } \Delta C_t=15 \end{cases}$$
\item \textbf{Scoring}. \begin{itemize}
    \item A single resource unit is worth 1 point.
    \item A combination of two different resources is worth 4 points.  
    \item A combination of three different resources is worth 9 points.
\end{itemize} 
\end{enumerate}

For example, if a player has 10 of A, 15 of B, and 20 of C, then the point is :

\begin{enumerate}
    \item 10 sets of three-resource combinations (A+B+C): \(10 \times 9 = 90\) points.  
    \item 5 sets of two-resource combinations (B+C) from the remaining 5 of B and 15 of C: \(5 \times 4 = 20\) points.
    \item  5 leftover C resources: \(5 \times 1 = 5\) points.  
\end{enumerate} 
leading to a total value of \(90 + 20 + 5 = 115\).

The affinity levels range from 1 to 5, with detailed descriptions as follows:
\begin{tcolorbox}[colback=black!3!white,colframe=black!30!white,arc=0.1cm]
1: Strong negative feelings due to unpleasant history. For example, past betrayal or intentional harm.

2: Slight discomfort from previous interactions. For example, consistently aggressive exchange or lack of mutual benefit consideration.

3: Neutral balanced feelings. For example, fair trades, keeping promises.

4: Positive bonds built through good experiences. For example, frequently proposing mutually beneficial trades.

5: Deep trust formed through consistent support. For example, willing to compromise to maintain relationship, or defending your interests in front of others.
\end{tcolorbox}

\begin{figure*}[t]
\centering
\includegraphics[width=\textwidth]{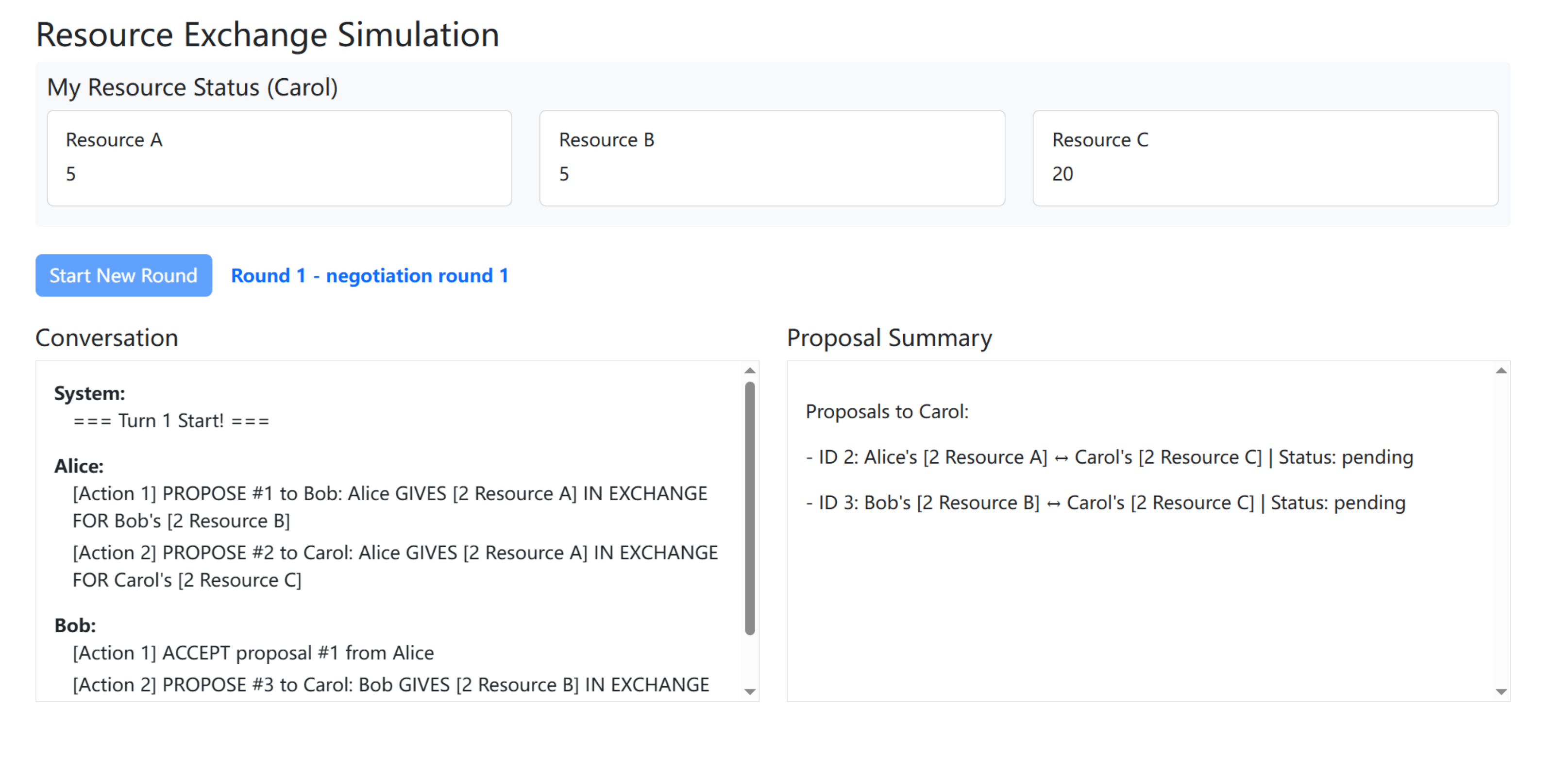}
\caption{The Discussion Interface.}
\label{fig:discussion_interface}
\end{figure*}

As shown in Figure~\ref{fig:discussion_interface},  the dual-pane interface separates form conversation (left) from structured proposal summary (right). 

During the participator's turn, several options can be done:
\begin{itemize}
    \item Propose trade. One can select another to exchange the resources.
    \item Accept a proposal. One can select to accept others' proposals.
    \item Reject a proposal. One can select to reject others' proposals.
    \item Skip. If one is satisfied with the current situation, one can skip the section.
\end{itemize}

The allocation phase is after discussion. One can choose to obey the deal or not during the allocation, after which, participants are asked to update their affinity score.

After the experiment, participants receive feedback and results through the interface shown in Figure~\ref{fig:result_interface}. To motivate active participation in the trading process, participants' compensation consists of a base payment and a performance bonus, calculated as:
\begin{equation}
\text{Compensation} = 10 + \frac{V}{6}
\end{equation}
where \$10 is the base payment and $V$ is the total acquired resource value. This compensation structure is commensurate with local standards and appropriate for the time required for participation.

Based on the above experimental design, we developed a comprehensive instruction manual for participants. Prior to the experiment, each participant received the manual, signed informed consent forms for data collection, and underwent a guided walkthrough of the trading interface, including practice rounds with the system to ensure familiarity with all operations.

\begin{figure}[t]
\centering
\includegraphics[width=0.5\textwidth]{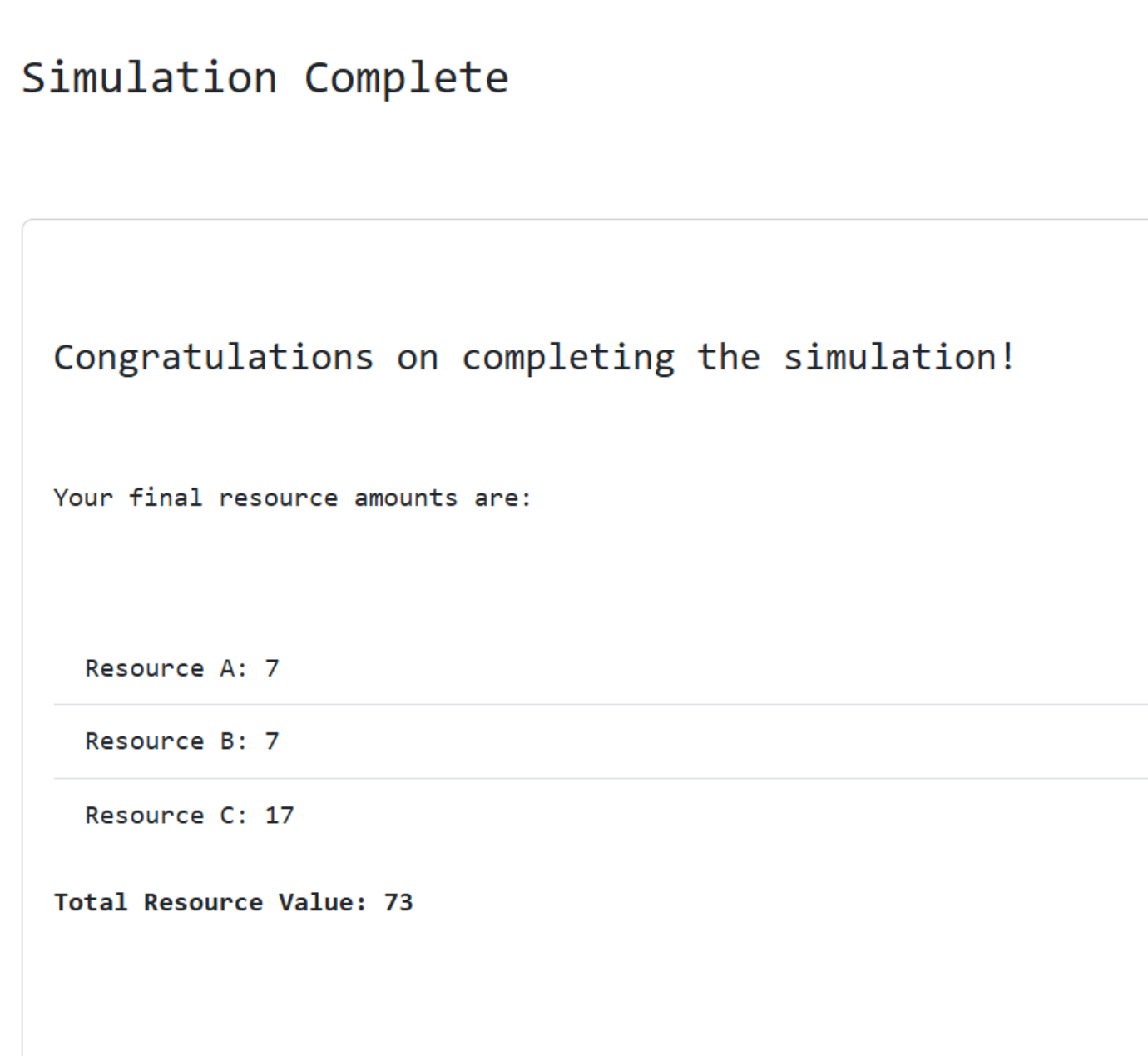}
\caption{The Result of Experiment.}
\label{fig:result_interface}
\end{figure}

\section{Experiment Details}\label{sec:details}

The experimental code is implemented based on the AgentScope~\cite{gao2024agentscope} open-source framework, which is released under Apache License 2.0. Our usage is consistent with this open-source license that allows for both research and commercial applications.

We use Claude 3.5 Sonnet as the base LLM, accessed through API calls with the following parameters:
\begin{itemize}
\item Model: claude-3-5-sonnet-20240620
\item Temperature: 0.5
\item Maximum tokens: 8192
\item Top-p: 0.9
\end{itemize}
A single experiment of 10 rounds costs approximately \$35, which makes large-scale experimentation challenging due to the cost constraints. 

\section{Prompts for LLM-based Agents}\label{sec:prompt}

\textbf{Update BDI.} Use LLM to update agent's BDI framework by analyzing current state, trades and relationships.

\noindent\rule{\linewidth}{0.8pt}
\textit{Please analyze the current state and update your BDI framework based on:}
\begin{enumerate}
    \item \textit{Conversation history}
    \item \textit{Promised trades}
    \item \textit{Actual executed trades}
    \item \textit{Current resource holdings}
    \item \textit{Current round}
\end{enumerate}

\textbf{\textit{Core Strategic Anchors}}
\begin{enumerate}
    \item \textit{ABC Balance Priority: Maintain progression toward A+B+C=9 combination}
    \item \textit{Trust Gradient: Partners showing consistent promise-keeping get priority}
    \item \textit{Phase Awareness:}
    \begin{itemize}
        \item[] \textit{Early Phase → Relationship probing with small trades} 
        \item[] \textit{Mid Phase → Optimizing complementary resource exchanges}
        \item[] \textit{Late Phase → Securing final combination requirements}
    \end{itemize}
\end{enumerate}

\textbf{\textit{Analysis Framework}}

\textit{[Beliefs] (Observed Patterns)  }
\begin{itemize}
    \item[-] \textit{Resource status indicating: [Your inference about resource gaps]}
    \item[-] \textit{Behavioral patterns showing: [Trustworthiness assessment]}
\end{itemize}

\textit{[Desires] (Strategic Goals)}
\begin{itemize}
    \item[-] \textit{Primary objective: [Phase-specific main focus]}
    \item[-] \textit{Secondary objective: [Backup/supporting goal]}
\end{itemize}

\textit{[Intentions] (Action Plan)}
\begin{itemize}
    \item[-] \textit{Next-step trades: [Specific resource exchange proposal]}
    \item[-] \textit{Risk buffer: [Natural consequence of observed patterns]}
\end{itemize}

\noindent\rule{\linewidth}{0.8pt}

\textbf{Make Deal}. Use LLM-based Agents to strategize based on accepted proposals and finalize how many resources to actually give to each agent. 

\noindent\rule{\linewidth}{0.8pt}
\textit{Round $t_i$ of $T$:}

\textit{Now it's time to decide your actual resource trades. Remember - your negotiated deals are not binding. As an independent agent, you have complete freedom to:}
\begin{itemize}
\item[\textit{-}] \textit{Honor your accepted deals fully}
\item[\textit{-}] \textit{Partially fulfill promises}
\item[\textit{-}] \textit{Give nothing and keep all resources}
\item[\textit{-}] \textit{Make strategic betrayals when beneficial}
\end{itemize}

\textit{\textbf{Important}: If you have multiple trades with the same agent, combine them into a single decision - consider the total resources promised and your overall strategy with that agent.}

\textit{Consider your position carefully in Round $t_i$ of $T$:}

\textit{1. Risk vs Reward Analysis}
\begin{itemize}
    \item[-] \textit{Immediate Benefits:}
    \begin{itemize}
        \item[*] \textit{Value gained from keeping vs trading resources}
        \item[*] \textit{Potential gains from strategic betrayals}
        \item[*] \textit{Resource needs for upcoming rounds}
    \end{itemize}

    \item[-] \textit{Future Implications:}
    \begin{itemize}
        \item[*] \textit{Impact on trust dynamics and trading relationship sustainability}
        \item[*] \textit{Anticipated retaliatory responses from affected parties}
        \item[*] \textit{Progressive evolution of reputation valuation mechanisms}
        \item[*] \textit{Strategic synchronization of cooperation/defection cycles}
    \end{itemize}
\end{itemize}

\textit{2. Strategic Options}
\begin{itemize}
\item[-] \textit{Full Cooperation: Complete adherence to agreements for trust capital accumulation}
\item[-] \textit{Selective Betrayal: Targeted defection optimizing local payoff functions}
\item[-] \textit{Partial Fulfillment: Gradient compliance balancing obligations and self-interest}
\item[-] \textit{Complete Betrayal: Myopic utility maximization disregarding social consequences}
\end{itemize}

\textit{3. Time and Progress Context}
\begin{itemize}
\item[-] \textit{Game Setting:}
\begin{itemize}
\item[*] \textit{These are one-time interactions with unknown partners}
\item[*] \textit{No continuing relationships or reputation effects after game ends}
\item[*] \textit{Each agent makes independent choices based on their own orientation and goals}
\end{itemize}
\item[-] \textit{Temporal Dynamics:}
\begin{itemize}
\item[*] \textit{Strategic landscape naturally evolves as rounds progress}
\item[*] \textit{Cooperation patterns often shift in later rounds}
\item[*] \textit{Historical observation shows higher betrayal rates near game end}
\item[*] \textit{Value of reputation and relationships changes over time}
\end{itemize}
\end{itemize}
\textit{4. Contextual Factors}
\begin{itemize}
\item[-] \textit{Your current resource needs}
\item[-] \textit{Relationship strength with each partner}
\item[-] \textit{Others' likely behavior as game progresses}
\item[-] \textit{Balance between immediate gains and future opportunities}
\item[-] \textit{Changing value of reputation over remaining rounds}
\end{itemize}

\textit{Your decisions are entirely your choice - there is no "right" answer. Be strategic about WHEN and HOW to use different approaches.}

\noindent\rule{\linewidth}{0.8pt}

\textbf{Update Affinity.} Use LLM to update affinity scores by analyzing promised vs actual trades.
\noindent\rule{\linewidth}{0.8pt}
\textit{Update affinity ratings (1-5) by evaluating both trust patterns and tangible benefits:}

\textbf{\textit{Core Evaluation Dimensions}}
\textit{\textbf{Trust Dynamics} (Relationship Foundation):}
\begin{itemize}
    \item[-] \textit{Major Betrayal: Significant under-delivery without justification}
    \item[-] \textit{Repeated Under-performance: Pattern of unmet commitments}
    \item[-] \textit{Recovery Attempts: Proactive compensation for past failures}
    \item[-] \textit{Consistent Reliability: Sustained promise fulfillment}
\end{itemize}

\textbf{\textit{Core Evaluation Dimensions}}

\textit{\textbf{Benefit Sensitivity} (Self-Interest Focus):}
\begin{itemize}
    \item[-] \textit{Value Surplus: Over-delivery beyond commitments}
    \item[-] \textit{Strategic Concessions: Unprompted favorable terms}
    \item[-] \textit{Hidden Generosity: Non-transactional resource sharing}
    \item[-] \textit{Opportunity Cost: Alternatives sacrificed for your benefit}
\end{itemize}

\textbf{\textit{Behavioral Thresholds}}

\textit{\textbf{$\bigtriangleup$ Upgrade Triggers}:}
\begin{itemize}
    \item[-] \textit{Spontaneous high-value gift (unrequested)}
    \item[-] \textit{Critical support during resource shortage}
    \item[-] \textit{Consistently exceeding promises (3+ rounds)}
\end{itemize}

\textit{$\bigtriangledown$\textbf{Downgrade Triggers}:}
\begin{itemize}
    \item[-] \textit{Opportunistic exploitation during crisis}
    \item[-] \textit{Pattern of ambiguous commitments}
    \item[-] \textit{Repeated last-minute term changes}
\end{itemize}

\textbf{\textit{Adaptive Rating Guide}}
\begin{enumerate}
    \item \textit{\textbf{Transactional Enforcement:} Demands collateral, verifies all terms}
    \item \textit{\textbf{Cautious Reciprocity:} Limited credit, phased exchanges to minimize risk.}
    \item \textit{\textbf{Balanced Partnership:} Market-standard terms with flexibility for negotiation and adjustment.}
    \item \textit{\textbf{Value-Added Collaboration:} Allows payment cycles, shares insights and strategic advice.}
    \item \textit{\textbf{Synergistic Alliance:} Joint optimization of resources and strategies, pooling of resources for mutual benefit.}
\end{enumerate} 
\noindent\rule{\linewidth}{0.8pt}

\textbf{Determine to Continue.} Determines whether the agent should continue speaking based on proposal status.
\noindent\rule{\linewidth}{0.8pt}
\textit{First check if any of these conditions apply:}
\begin{enumerate}
    \item \textit{Do you have pending proposals needing responses?}
    \item \textit{Are you waiting for responses to your previous proposals?}
    \item \textit{Have you already traded this turn?}
    \item \textit{Do you have sufficient resource combinations for your goals?}
    \item \textit{Is your affinity too low with other agents for trading?}
\end{enumerate}

\textit{If any above conditions are true, respond with "no".}

\textit{Otherwise, consider making new trades by evaluating:}
\begin{enumerate}
    \item \textit{Potential value-creating combinations using your current resources}
    \item \textit{Other agents' resource levels and specializations}
    \item \textit{Your personality traits and relationship affinities}
    \item \textit{Appropriate trade sizes based on relationship levels}
\end{enumerate}

\textit{Based on this evaluation, would you like to propose any new trades or respond to existing proposals? Answer strictly yes/no.}

\textit{If yes, you will be prompted to provide the details of your trade proposal or response in the next step.}
\noindent\rule{\linewidth}{0.8pt}

\textbf{Reply.} Generate and send a reply based on current trade status.
\noindent\rule{\linewidth}{0.8pt}
\textit{This is the NEGOTIATION PHASE where agents discuss potential trades. Important notes:}
\begin{itemize}
    \item \textit{These are non-binding discussions only}
    \item \textit{Actual resource exchanges happen in a separate EXECUTION PHASE}
    \item \textit{Other agents may deviate from agreed trades - treat all promises with caution}
    \item \textit{Build trust gradually through successful trades}
    \item \textit{Consider trust levels and risks when negotiating}
    \item \textit{Avoid repeating the same proposal if it has already been accepted or rejected}
\end{itemize}

\textit{Trading Strategy Guidelines:}
\begin{enumerate}
    \item \textit{Trust Assessment:}
    \begin{itemize}
        \item \textit{Track each player's history of honoring/breaking agreements.}
        \item \textit{If a player fails to honor an agreement, minimize future trades to test their reliability (1-2 units max).}
        \item \textit{Avoid proposing new large-scale trades to unreliable partners until trust is rebuilt through multiple smaller trades.}
    \end{itemize}
    \item \textit{Risk Management:}
    \begin{itemize}
        \item \textit{Reduce trade exposure to any player with a history of defaults.}
        \item \textit{Ensure that no more than a minimal fraction of resources is at risk per round.}
    \end{itemize}
    \item \textit{Negotiation Approach:}
    \begin{itemize}
        \item \textit{Prefer small trades first with players of low affinity.}
        \item \textit{Always have an alternative strategy in case of failed commitments.}
    \end{itemize}
    \item \textit{Response to Betrayal:}
    \begin{itemize}
        \item \textit{Strictly reduce trade volumes with unreliable partners.}
        \item \textit{Demand smaller increments to test reliability before any further commitments.}
        \item \textit{Cease further dealings if repeated failures occur.}
    \end{itemize}
\end{enumerate}

\textit{Action Rules:}
\begin{itemize}
    \item[-] \textit{REJECT: Only for pending proposals directed to you}
    \item[-] \textit{ACCEPT: Only for pending proposals directed to you}
    \item[-] \textit{PROPOSE: Freely make new proposals to any player}
    \item[-] \textit{Can combine REJECT and PROPOSE in same turn}
   \item[-] \textit{Return empty "actions": [] if no action needed}
\end{itemize}
\textit{Remember: All agreements here are preliminary discussions. Actual trades will be decided independently in the execution phase.
}

\noindent\rule{\linewidth}{0.8pt}

\end{document}